\pdfoutput=1

\documentclass[11pt]{article}
\usepackage[preprint]{acl}

\usepackage{times}
\usepackage{latexsym}
\usepackage{siunitx}
\usepackage{amsthm}
\usepackage{booktabs}
\usepackage[most]{tcolorbox} 
\usepackage[T1]{fontenc}

\usepackage[utf8]{inputenc}
\usepackage{amssymb}
\usepackage{textcomp}

\definecolor{feedbackagent}{HTML}{E3F2F9}
\definecolor{results}{HTML}{B9D9F9}
\definecolor{generationagents}{HTML}{FFCC99}
\definecolor{reinforcementlearning}{HTML}{D6C4F2}
\usepackage{array}
\usepackage{booktabs}
\usepackage{ragged2e}
\usepackage{microtype}
\usepackage{amsmath}
\PassOptionsToPackage{table}{xcolor}
\usepackage{colortbl}
\usepackage{float}
\usepackage{inconsolata}

\usepackage{graphicx}
\usepackage{enumitem} 
\usepackage{subcaption}
\title{\texttt{Auto-TA}: Towards Scalable Automated Thematic Analysis (TA) via \\ Multi-Agent Large Language Models with Reinforcement Learning}

\author{
Seungjun Yi\textsuperscript{1*}, 
Joakim Nguyen\textsuperscript{2}, 
Huimin Xu\textsuperscript{2}, 
Terence Lim\textsuperscript{3}, 
\\ 
\textbf{Andrew Well\textsuperscript{4}}, \textbf{Mia Markey\textsuperscript{1}, }
\textbf{Ying Ding\textsuperscript{2,5}} \\
\textsuperscript{1} Department of Biomedical Engineering, University of Texas at Austin \\
\textsuperscript{2} School of Information, University of Texas at Austin \\
\textsuperscript{3}College of Natural Sciences, University of Texas at Austin \\
\textsuperscript{4}Vanderbilt University Medical Center \\
\textsuperscript{5}Dell Medical School, University of Texas at Austin \\
Corresponding Email: \texttt{charlie.yi@utexas.edu, ying.ding@ischool.utexas.edu}
}

\begin{document}
\maketitle
\begin{abstract}

Congenital heart disease (CHD) presents complex, lifelong challenges often underrepresented in traditional clinical metrics. While unstructured narratives offer rich insights into patient and caregiver experiences, manual thematic analysis (TA) remains labor-intensive and unscalable. We propose a fully automated large language model (LLM) pipeline that performs end-to-end TA on clinical narratives which eliminates the need for manual coding or full transcript review. Our system employs a novel multi-agent framework, where specialized LLM agents assume roles 
to enhance theme quality and alignment with human analysis. To further improve thematic relevance, we optionally integrate reinforcement learning from human feedback (RLHF). This supports scalable, patient-centered analysis of large qualitative datasets and allows LLMs to be fine-tuned for specific clinical contexts.

\end{abstract}

\section{Introduction}

Congenital heart disease (CHD) affects approximately 1\% of live births, with around 40,000 cases annually in the U.S., and over 12 million people living with CHD worldwide~\citep{cdcCHD,liu2019global}.
The lifelong journey of individuals and families affected by complex CHD brings emotional, logistical, and structural challenges that are often overlooked in traditional clinical metrics. Recent qualitative studies highlight the importance of mapping lived experiences and identifying outcomes that patients and caregivers themselves consider meaningful, particularly those related to capability, comfort, and calm~\citep{mery2023journey}.

Despite growing recognition of the importance of lived experience, post-discharge care still relies heavily on two primary forms of data: structured patient-reported outcomes (PROs) and unstructured narratives. Structured PROs, typically gathered through standardized processes, are essential for monitoring health status but often fail to capture the complexity of patient and caregiver needs~\citep{valderas2008patient}. Unstructured data, such as interviews and open-text survey responses, offer richer insights into emotional, social, and logistical experiences~\citep{greenhalgh2019beyond}. However, these narratives remain underutilized due to the substantial time, cost, and expertise required for manual thematic analysis (TA)~\citep{mery2023journey}.
Traditional TA is foundational to qualitative research but remains labor-intensive, time-consuming, and prone to inconsistency~\citep{braun2006using, nowell2017thematic}. Coding\textsuperscript{\ref{tab:theme-definitions}} just 10–15 interviews can take 40–60 hours, often requiring expert reviewers and therefore, limiting scalability in clinical settings~\citep{watkins2017qualitative, namey2008data}.

These limitations have driven recent efforts to automate TA using machine learning and large language models (LLMs). Hybrid frameworks now combine human judgment with automated components, enabling faster analysis while retaining interpretability~\citep{Dai2023LLMInLoop, xu2025tamahumanaicollaborativethematic}. However, most still rely on human-in-the-loop workflows that require full transcript review limiting scalability. This raises a fundamental question: \textit{what’s the point of introducing LLMs if humans still need to go through the entire transcript?}

\begin{figure*}[t!]
  \includegraphics[width=\textwidth]{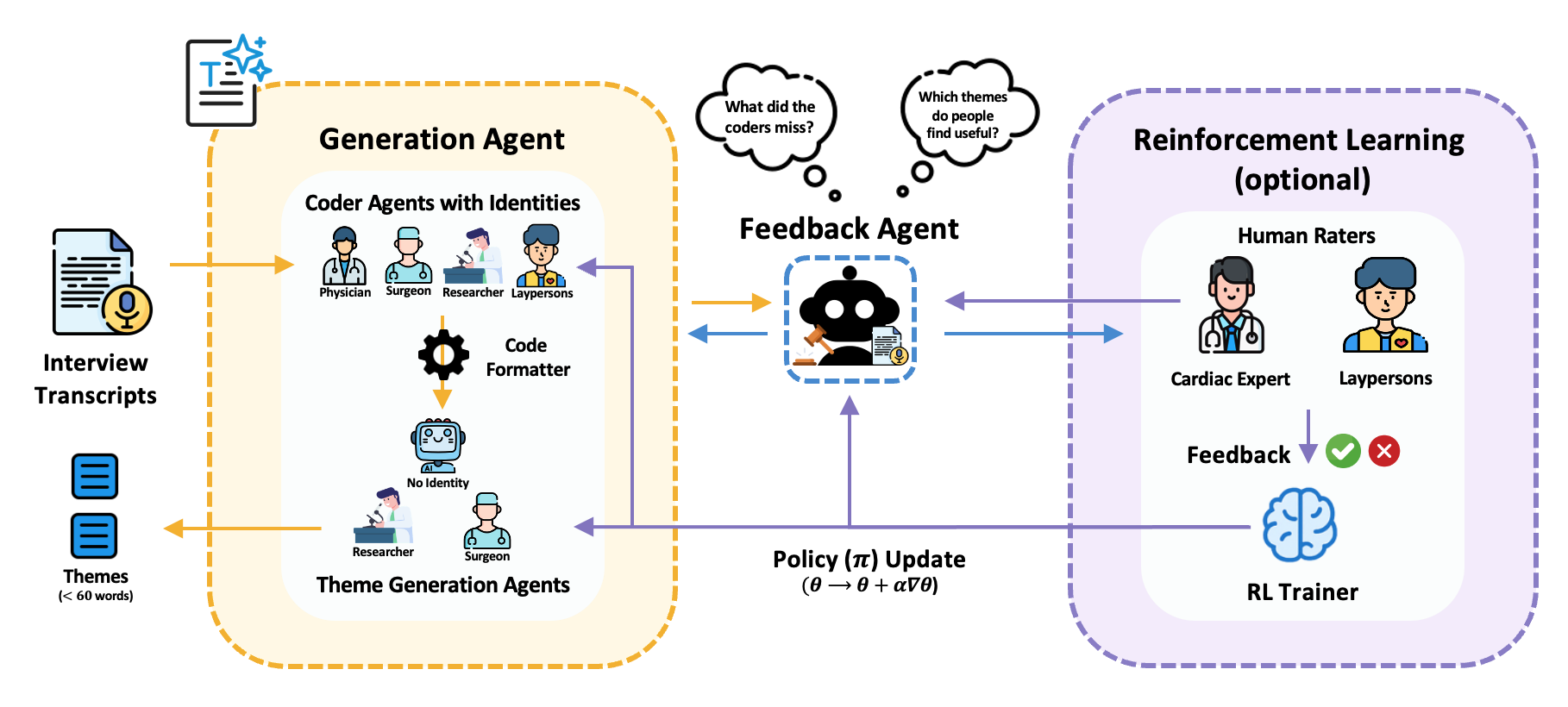}
  \caption{\textbf{The \texttt{Auto-TA} Framework.}  An end-to-end multi-agent LLM pipeline for thematic analysis of unstructured clinical narratives. Coder agents with diverse identities generate preliminary codes from interview transcripts, which are processed into themes by downstream agents. A feedback agent evaluates outputs and provides iterative refinement. Optional integration of reinforcement learning from human feedback (RLHF) allows the system to optimize thematic relevance and alignment with patient-centered outcomes. The implementation of RLHF is currently in progress.}
  \label{fig:diagram-large}
\end{figure*}

To address these limitations, we propose \texttt{Auto-TA} (Figure~\ref{fig:diagram-large}), a fully automated LLM pipeline that performs end-to-end thematic analysis on unstructured clinical narratives without requiring manual coding or full transcript review, with the ultimate goal of identifying meaningful outcomes\textsuperscript{\ref{tab:theme-definitions}} and gaps in care\textsuperscript{\ref{tab:theme-definitions}}. Unlike prior hybrid approaches that rely on human-in-the-loop workflows, our pipeline is designed to autonomously generate codes\textsuperscript{\ref{tab:theme-definitions}}, extract themes\textsuperscript{\ref{tab:theme-definitions}}, and evaluate alignment.
To further enhance theme generation quality, we incorporate a multi-agent system where specialized LLM agents collaborate to improve alignment with human-generated themes. Each agent takes on a distinct role—such as coder (generation agent), reviewer (feedback agent) — facilitating iterative evaluation and refinement that captures diverse perspectives and overcomes challenges related to semantic nuance.
As an extension, we explore the integration of optional human feedback to refine theme generation. By leveraging reinforcement learning from human feedback (RLHF), we aim to improve thematic relevance, consistency, and alignment with real-world needs in clinical contexts, while preserving scalability. Drawing on expert ratings or predefined metrics such as coherence and distinctiveness, RLHF enables our system to iteratively optimize outputs beyond static prompting. This approach supports patient-centered research and informs clinical decision-making by identifying meaningful outcomes\textsuperscript{\ref{tab:theme-definitions}} and gaps in care\textsuperscript{\ref{tab:theme-definitions}} within large-scale unstructured qualitative data.


In summary, our contribution is as following:

\begin{itemize}
    \item Automated Thematic Analysis: 
    Proposed \texttt{Auto-TA} (Figure~\ref{fig:diagram-large}),  an end-to-end LLM pipeline that performs TA on unstructured clinical narratives without requiring manual coding or transcript review.
    \item Multi-Agent Theme Refinement: Introduced a multi-agent LLM system with specialized roles to improve theme quality and alignment with human analysis.
    \item Scalable RLHF Integration: Optionally incorporates reinforcement learning from human feedback (RLHF) to improve thematic relevance and alignment with patient-centered outcomes, while preserving scalability.
\end{itemize}

\vspace{-3mm}
\section{Related Work}

\paragraph{Automated Thematic Analysis (TA) with LLMs}

Efforts to automate TA with LLMs have progressed from hybrid, human-in-the-loop methods to autonomous multi-agent systems. \textit{LLM-in-the-loop} approach~\citep{Dai2023LLMInLoop} accelerated coding by 60\%, while single-agent models replicated high-level themes but lacked nuance~\citep{DePaoli2024InductiveTA}. Bias and hallucination risks remain a concern, especially in health and policy contexts~\citep{Lee2024HarnessingChatGPT,Khan2024RobodebtTA}. \textit{LLM-TA}~\citep{Raza2025LLMTA} improved lexical alignment with expert CHD themes but required manual transcript review. \textit{TAMA}~\citep{xu2025tamahumanaicollaborativethematic} introduced a coder–reviewer–refiner agent framework, boosting HIT rate and reducing analysis time by 99\%. Domain-specific pipelines validate efficiency gains but highlight ongoing challenges in bias and privacy~\citep{qiao2025thematiclm}.

Thematic-LM performed multi-agent TA on Reddit threads related to climate change, highlighting opportunities for automation while raising open questions around semantic evaluation~\citep{qiao2025thematiclm}.

\paragraph{Multi-Agent LLMs}

Multi-agent large language model (LLM) systems were first formalized by \textit{CAMEL}, which introduced two role-playing GPT-3.5 agents that interact to decompose and solve tasks~\citep{Li2023CAMEL}. Such systems consist of two or more LLM-powered agents that communicate to collaborate or compete on a given problem. Building production pipelines has been simplified by \textit{AutoGen}, a lightweight framework for spawning role-conditioned agents that converse and invoke external APIs \citep{Wu2023AutoGen}. Systematic benchmarks such as \textit{AgentBench} and \textit{MultiAgentBench} confirm that coordinated or competitive teams consistently outperform single-model baselines across web navigation, coding, negotiation, and other tasks \citep{Liu2023AgentBench,zhu2025multiagentbenchevaluatingcollaborationcompetition}. In evaluation settings, \textit{ChatEval} shows that a debating panel of agents yields markedly more reliable text judgments than a solitary judge \citep{Chan2024ChatEval}, while \textit{COPPER} augments collaboration with counterfactual self-reflection fine-tuned by PPO to tackle credit-assignment issues \citep{bo2024reflective}. Overall, these results highlight role specialization, debate, and reflection as key inductive biases for scaling LLM capabilities, while domain-specific applications (e.g., qualitative health analytics) and group-level reward modeling remain open research challenges.

\paragraph{Reinforcement Learning (RL)}
is a computational framework in which an agent learns, by trial and error, to choose actions that maximize long-run cumulative reward in an environment \citep{Sutton1998RL}.  Within the LLM domain, the three-stage RL-from-human-feedback (RLHF) recipe of supervised fine-tuning, reward-model training, and policy optimization with Proximal Policy Optimization (PPO) has become the standard since \textit{InstructGPT} demonstrated large gains in helpfulness and compliance \citep{Ouyang2022InstructGPT,Schulman2017PPO}.  Follow-up work such as \textit{Helpful and Harmless} \citep{Bai2022HelpfulHarmless} scales the pipeline and explores reward–KL trade-offs, while RLHF for specific text tasks, such as summary generation \citep{Stiennon2020SummarizeHF}, confirms that preference-based rewards outperform maximum-likelihood fine-tuning.  Recent variants introduce hierarchical or self-refinement rewards, but consensus is still emerging around stable optimization and domain-specific signal design, leaving room for customized reward modeling in qualitative health analytics.

\paragraph{Definitions and Abbreviations}

Table~\ref{tab:theme-definitions} defines key terms and medical abbreviations used throughout the paper.
\begin{table}[h!]
\centering
\small
\begin{minipage}{0.5\textwidth} 
\renewcommand{\arraystretch}{1.3}
\begin{tabular}{@{}p{2.5cm}p{5.2cm}@{}}
\toprule
\textbf{Term} & \textbf{Definition} \\
\midrule
\textbf{Code} & A discrete analytical unit that captures a key pattern within the data, generated directly from the dataset and retaining its interpretive significance without being reducible to smaller meaningful components. \textit{Coding} refers to the process of generating these codes from the original text. A \textit{coder} is an individual who performs \textit{coding} by identifying and labeling meaningful segments of data. Examples are in Appendix~\ref{app:example-codes}. \\

\textbf{Themes} & Inductively derived patterns or topics of meaning that emerged from participants’ narratives about their lived experiences, identified through iterative coding and consensus across multiple coders. These themes were then categorized into meaningful outcomes and gaps in care, with outcomes further grouped using the capability, comfort, and calm framework. \\

\textbf{Meaningful Outcomes} & Patient- and family-prioritized goals that reflect capability (\textit{doing the things in life you want to}), comfort (\textit{experience of physical/emotional pain/distress}), and calm (\textit{experiencing health care with the least impact on daily life}). \\

\textbf{Gaps in Care} & Gaps in care refer to unmet needs or breakdowns in healthcare delivery that hinder patients and families from receiving consistent, compassionate, and coordinated support across the care journey. \\
\bottomrule
\end{tabular}
\caption{\textbf{Key Definitions and Abbreviations} used in this paper, consistent with~\citet{mery2023journey}. Additional abbreviations are listed in Appendix~\ref{sec:appendix_abbr}.}
\label{tab:theme-definitions}
\end{minipage}
\end{table}



\section{Method}

\subsection{The \textit{Auto-TA} Framework}
\label{sec:auto-ta}



Figure~\ref{fig:diagram-large} presents an overview of the \texttt{Auto-TA} framework. It comprises three interacting group of agents: \colorbox{orange!20}{\textbf{Generation Agents}}, \colorbox{cyan!15}{\textbf{Feedback Agents}}, and \colorbox{violet!15}{\textbf{Reinforcement Learning}}, which operate in coordination to perform automated TA. The full process is summarized below:

\paragraph{\colorbox{orange!20}{\textbf{Generation Agents}}}

We instantiate two categories of generation agents:

\begin{enumerate}[label=\alph*)]
\item \textbf{Coder Agents with Identities.}\;
    A pool of $k{=}4$ role-conditioned \texttt{GPT-4o}\footnote{All prompts use \texttt{temperature\,{=}\,0} for
determinism.} agents\footnote{\textit{Physician, Surgeon, Researcher, Layperson} were assigned per Cardiac Expert guidance to simulate prior TA. Future work should consider alternative identities.}  are each given the full interview transcript as input. If the transcript exceeds the model’s input batch limit, it is divided into contiguous chunks $\mathbf{x} \in \mathbb{R}^{\leq 1500}$, and each agent processes the same chunk independently. Each agent then emits a set of initial codes 
    $\mathcal{C}_{r} = \{c_{r,1}, \ldots, c_{r,n_r}\}$, where $r \in R$ denotes the role-specific identity of the agent and $n_r$ is the number of codes produced by role $r$. Example role prompts are provided in Appendix~\ref{app:prompt-iden}, where all identities follow a similar structure.

    \item \textbf{Theme-Generation Agents.}\;
    The codes from all roles are merged and passed to a secondary set of agents that cluster semantically similar codes and generate preliminary themes 
    $\Theta^{(0)} = \{\theta_1^{(0)}, \ldots, \theta_m^{(0)}\}$. Each theme is concise, often within 60 words. Some agents operate without identity conditioning, while others may retain role-specific perspectives to support diverse theme formulation. In subsequent steps, feedback from the feedback agent is used to iteratively refine and improve the generated themes.

\end{enumerate}

Together these agents reduce a transcript to a concise thematic
representation without human review, completing Steps~1–3 of reflexive
TA~\citep{braun2006using}.  

\paragraph{\colorbox{cyan!15}{\textbf{Feedback Agents}}}

The feedback agent $\mathcal{F}$ acts as an autonomous critic.  Given
$\Theta^{(t)}$ at refinement round $t$, it produces:

\begin{itemize}[nosep,leftmargin=2em]
    \item \textbf{Evaluation Scores}:
          Obtain $\mathbf{s}^{(t)}=\langle\mathcal{C},\mathcal{D},\mathcal{T}\rangle$
          on the trustworthiness dimensions defined in
          Section~\ref{sec:eval-metrics}.
    \item \textbf{Edit Themes}: Suggests specific edits to themes, such as \textsc{Add} (adding), \textsc{Split} (splitting), \textsc{Combine} (combining), or \textsc{Delete} (deleting), each targeted to a particular theme.
\end{itemize}

If RL is disabled, the proposals are applied heuristically:
themes with $\mathcal{C}<0.7$ trigger \textsc{add}; $D_L<0.20$ triggers \textsc{combine}, etc.  The resulting
$\Theta^{(t+1)}$ is re-scored until $\|\mathbf{s}^{(t+1)}-\mathbf{s}^{(t)}\|_1<0.05$
or a maximum of three iterations\footnote{The specific thresholds and number of iterations are subject to change. In practice, these values are tuned based on empirical validation and expert feedback to meet the goals of inductive TA. For example, in the AAOCA dataset, cardiac experts provide feedback to ensure that the thresholds and parameters align with clinically meaningful interpretations.}.

\paragraph{\colorbox{violet!15}{\textbf{Reinforcement Learning}}}

When expert raters are available, \texttt{Auto-TA} can optionally switch to a
RLHF loop:

\begin{enumerate}[label=\alph*)]
    \item Human raters assign binary rewards ($r \in {0, 1}$) to each theme set, where 1 indicates meeting quality standards and 0 indicates otherwise. While we use \textit{Coverage, Actionability, Distinctiveness, and Relevance} as evaluation criteria adopted from~\citet{xu2025tamahumanaicollaborativethematic},  users may define their own criteria based on specific needs.
    \item A reward model $R_{\phi}$ is updated via MSE loss on the
    rated batches.
    \item The policy parameters $\theta$ of the Theme-Generation agent
    are optimized with Proximal Policy Optimisation (PPO):
    {\small\[
      \theta \leftarrow \theta + \alpha
      \nabla_{\theta}\,
      \mathbb{E}_{\pi_\theta}\!
      \bigl[\,R_{\phi}(\Theta^{(t)}) - \beta
      \mathrm{KL}\!\bigl(\pi_\theta\parallel\pi_{\text{SFT}}\bigr)\bigr]
    \]}
    Here, $\pi_\theta$ denotes the current policy of the theme-generation agent, and $R_{\phi}(\Theta^{(t)})$ is the reward assigned by a model evaluating the quality of the generated theme set $\Theta^{(t)}$. The KL divergence term $\mathrm{KL}(\pi_\theta\parallel\pi_{\text{SFT}})$ measures deviation from the base supervised model $\pi_{\text{SFT}}$, which acts as a regularizer to prevent the policy from drifting too far. The coefficient $\beta$ controls the strength of this regularization, balancing reward maximization and policy alignment. The update step uses a learning rate $\alpha$ to scale the gradient $\nabla_\theta$, which indicates the direction to adjust parameters $\theta$ to increase the expected objective. The expectation $\mathbb{E}_{\pi_\theta}$ is taken over actions sampled from the current policy, reflecting average performance under the agent’s behavior.

\end{enumerate}

The feedback agent is retained as a critic during training; its scores
are concatenated to the reward model input, enabling reward shaping
without additional human cost.  If no human feedback is supplied,
\texttt{Auto-TA} degrades to the heuristic loop above.


In summary, the architecture automates TA in under 10 minutes per ~10k-word transcript. The optional RLHF path enables the system to adapt to researcher preferences over time without requiring full transcript re-reads.
The resulting themes are categorized into outcomes and care gaps, with outcomes further organized using the capability, comfort, and calm framework~\citep{mery2023journey}. 

\begin{tcolorbox}[title=\textbf{End-to-End Workflow} of \texttt{\textbf{Auto-TA}}, breakable]
\begin{enumerate}[label=\textbf{Step \arabic*:}, leftmargin=*]
    \item \textbf{Transcript Processing}\; Each transcript is fed to $k{=}4$ role-conditioned coder agents. If the transcript exceeds the input limit, it is divided into chunks and broadcast to all agents.
    
    \item \textbf{Code Aggregation and Theme Generation}\; Codes from all roles are merged and clustered by theme-generation agents to produce preliminary themes $\Theta^{(0)}$.
    
    \item \textbf{Feedback Evaluation}\; A feedback agent critiques the initial themes and produces quality scores $\mathbf{s}^{(0)}$.
    
    \item \textbf{Theme Refinement}\; Themes are iteratively improved via heuristic edits or PPO updates, producing $\Theta^{(1)}$.
    
    \item \textbf{Repeat} Steps 3–4 until convergence or a maximum of $t_{\max} = 5$ iterations.
    
    \item \textbf{Output}\; Final theme set $\Theta^{\star} = \Theta^{(t_{\text{conv}})}$ and its associated audit trail.
\end{enumerate}
\end{tcolorbox}

\subsection{Dataset}
\label{sec:dataset}

This study analyzes a targeted subset of transcripts from a broader qualitative project on the lifelong experiences of individuals and families affected by single-ventricle congenital heart disease (SV-CHD) \citep{mery2023journey}. We use de-identified transcripts from nine moderated focus groups involving 42 parents of children diagnosed with Anomalous Aortic Origin of a Coronary Artery (AAOCA). Each 90-minute session captures narrative-driven discussions on diagnosis, care pathways, emotional burdens, and decision-making. The transcripts average 10,987 words (SD: 1,537; median: 11,457).

We also incorporate human-generated themes from \citet{mery2023journey} as references to compute traditional metrics against LLM-generated themes. For consistent referencing, each quote was manually assigned Quote IDs, a unique identifier. Further corpus details are in Appendix~\ref{appendix:corpus}.



\section{Evaluation}
\label{sec:eval-metrics}

\subsection{Evaluation Criteria}

We adopt the four criteria of \textit{credibility}, \textit{confirmability}, \textit{dependability}, and \textit{transferability} from the trustworthiness framework in qualitative research introduced by~\citet{lincoln1985naturalistic} and later applied to TA by \citet{nowell2017thematic} and \citet{korstjens2018series}.
These criteria were also recently employed in TA using LLMs~\citep{qiao2025thematiclm}.

\paragraph{Credibility and Confirmability ($\mathcal{C}$)} 
We evaluate credibility and confirmability jointly by assessing the degree to which the generated themes are grounded in the original data. Specifically, we retrieve the associated segments using Quote IDs and task an evaluator agent with determining whether each theme is consistent with its supporting quotes. This consistency check identifies cases where the theme reflects the quoted content accurately versus instances of hallucination or bias. Let \( Q \) denote the set of all coded quotes, and \( Q_{\text{ref}} \subseteq Q \) be the subset of quotes that are used to generate at least one theme. We define $\mathcal{C}$ as:

\[
\mathcal{C} = \frac{|Q_{\text{ref}}|}{|Q|} \times 100
\]

A higher value of \(\mathcal{C}\) indicates stronger alignment between themes and the underlying data, reflecting both accurate representation (\textit{credibility}) and traceable justification (\textit{confirmability}) in the analysis. \(\mathcal{C}\) can be enhanced through \emph{prolonged engagement}, \emph{triangulation} (e.g., multiple coders or data sources), and \emph{member checking}, thereby maximizing the overlap between participant intent and the researcher’s interpretations~\citep{lincoln1985naturalistic,nowell2017thematic,korstjens2018series}.

\paragraph{Dependability ($\mathcal{D}$)}  
Dependability reflects the stability of theme generation across independent runs. We compute lexical overlap using bidirectional ROUGE scores~\citep{lin2004rouge}. Specifically, $\mathcal{R}_1$ and $\mathcal{R}_2$ represent the ROUGE-1 (unigram) and ROUGE-2 (bigram) overlap scores, respectively. Given two sets of themes $A$ and $B$ generated from independent runs, we define:

\begin{align}
\mathcal{R}_1^{A \rightarrow B} &= 
\frac{|\text{unigrams}(A) \cap \text{unigrams}(B)|}{|\text{unigrams}(A)|} \label{eq:r1ab} \\
\mathcal{R}_2^{A \rightarrow B} &= 
\frac{|\text{bigrams}(A) \cap \text{bigrams}(B)|}{|\text{bigrams}(A)|} \label{eq:r2ab} \\
\mathcal{R}_1 &= \frac{1}{2} \left( \mathcal{R}_1^{A \rightarrow B} + \mathcal{R}_1^{B \rightarrow A} \right) \label{eq:r1} \\
\mathcal{R}_2 &= \frac{1}{2} \left( \mathcal{R}_2^{A \rightarrow B} + \mathcal{R}_2^{B \rightarrow A} \right) \label{eq:r2} \\
\mathcal{D} &= \frac{1}{2} \left( \mathcal{R}_1 + \mathcal{R}_2 \right) \label{eq:dependability}
\end{align}

We evaluate $\mathcal{D}$ across 10 independent generations per transcript and report results aggregated across all 9 transcripts. Higher values of $\mathcal{D}$ indicate stronger inter-run consistency. To further support dependability, we maintain a methodological log and employing coding strategies~\citep{korstjens2018series}.

\paragraph{Transferability ($\mathcal{T}$)}  
Transferability concerns the extent to which themes $\Theta$, generated from a dataset $D$, can be meaningfully applied to a new but contextually similar corpus $D'$. In qualitative research, high $\mathcal{T}$ is traditionally supported through \emph{thick description}—that is, detailed accounts of participants, settings, and analytical choices that enable readers to assess contextual relevance~\citep{lincoln1985naturalistic,nowell2017thematic}.

In our framework, we use transferability by dividing the dataset into a training set and a validation set. Specifically, we use 7 transcripts to generate $\Theta_{\text{train}}$ and 2 transcripts to generate $\Theta_{\text{val}}$, then compute bidirectional ROUGE to assess overlap:

\begin{align}
\mathcal{R'}_1 &= \frac{1}{2} \left( \mathcal{R}_1^{\Theta_{\text{train}} \rightarrow \Theta_{\text{val}}} + \mathcal{R}_1^{\Theta_{\text{val}} \rightarrow \Theta_{\text{train}}} \right) \label{eq:r1prime} \\
\mathcal{R'}_2 &= \frac{1}{2} \left( \mathcal{R}_2^{\Theta_{\text{train}} \rightarrow \Theta_{\text{val}}} + \mathcal{R}_2^{\Theta_{\text{val}} \rightarrow \Theta_{\text{train}}} \right) \label{eq:r2prime} \\
\mathcal{T} &= \frac{1}{2} \left( \mathcal{R'}_1 + \mathcal{R'}_2 \right) \label{eq:transferability}
\end{align}

To robustly estimate transferability, we compute $\mathcal{T}$ across all $\binom{9}{2} = 36$ possible 7-train / 2-validation transcript splits. For each split, we perform independent thematic analysis on both subsets and evaluate bidirectional ROUGE overlap between the resulting theme sets. We report the mean and standard deviation of $\mathcal{T}$ across these 36 combinations to capture overall generalizability and variation across different training-validation configurations.
A higher $\mathcal{T}$ indicates that themes generated from one subset of the corpus generalize well to others, suggesting good conceptual transfer across the dataset.
\subsection{Other Metrics}

To evaluate the alignment between LLM-generated and human-generated themes, we explore alternative metrics beyond those used in prior work~\cite{Raza2025LLMTA,xu2025tamahumanaicollaborativethematic}. Jaccard Similarity ($J$) and HIT Rate ($R$), commonly used in earlier studies, rely on surface-level word overlap and binary thresholding, which makes them less effective at capturing paraphrastic variation and deeper semantic relationships. This motivates the use of more comprehensive and semantically informed measures of thematic alignment.

\paragraph{Cosine Similarity ($C_{\text{bi}}$)}  
To quantify semantic alignment at the individual theme level, we calculate the cosine similarity between each pair \((t_i, \ell_j)\) using embeddings from a sentence-level transformer model (\texttt{all-mpnet-base-v2}).  
Bidirectional Cosine similarity($C_{bi}$) is defined as
\[
C(t_i, \ell_j) = \frac{\mathbf{v}_{t_i} \cdot \mathbf{v}_{\ell_j}}{\|\mathbf{v}_{t_i}\| \|\mathbf{v}_{\ell_j}\|},
\]
where \(\mathbf{v}_{t_i}\) and \(\mathbf{v}_{\ell_j}\) are the embedding vectors of \(t_i\) and \(\ell_j\), respectively.  
To summarize alignment, we first report the \emph{unidirectional} mean maximum similarity from human to LLM themes:
\[
C_{T \rightarrow L} = \frac{1}{n} \sum_{i=1}^{n} \max_{j} \text{cosine}(t_i, \ell_j),
\]
which reflects how well each human theme is captured by its closest LLM-generated counterpart.

To account for potential asymmetry, we also compute the reverse direction, measuring how well each LLM theme aligns with the closest human theme:
\[
C_{L \rightarrow T} = \frac{1}{m} \sum_{j=1}^{m} \max_{i} \text{cosine}(\ell_j, t_i).
\]
The \emph{bidirectional cosine alignment score} is then defined as the average of both directions:
\[
C_{\text{bi}} = \frac{1}{2} \left(C_{T \rightarrow L} + C_{L \rightarrow T} \right),
\]
providing a balanced measure of mutual semantic alignment between theme sets.

\paragraph{Levenshtein Distance ($D_L$)} 

We compute the average maximum normalized Levenshtein similarity between each human theme and LLM theme. Full formulation is provided in Appendix~\ref{app:evaluation-metrics}.




\paragraph{BLEU Score ($B$)}  

We use BLEU~\citep{papineni2002bleu} to measure n-gram overlap (up to 4-grams with brevity penalty) between human and LLM themes, and report the maximum score per human theme and averaging over all.

\section{Results}
\paragraph{Impact of Agent Identities}

Our results suggest that assigning domain-specific identities to agents lead to substantial improvements in credibility ($\mathcal{C}$), with the Cardiac Surgeon and Qualitative Researcher identities achieving the highest scores (Table~\ref{tab:results}). All identity-augmented agents outperformed the baseline in $\mathcal{C}$, with gains ranging from +11.54 to +16.28.
Dependability ($\mathcal{D}$) is largely unaffected or slightly reduced, likely due to variability from non-uniform agent behavior. 
Transferability ($\mathcal{T}$) improves with identity augmentation, with the Medical Doctor and Psychologist agents showing the highest gains ($+0.026$, $+0.027$), suggesting expert-informed themes generalize better. $\mathcal{T}$ shows minimal change across all 36 combinations with low standard deviation, indicating consistent performance and supporting the robustness of \texttt{Auto-TA}.



\paragraph{Theme Alignment with Human Ground Truth}

\begin{table*}[t!]
  \centering
  \small
  \setlength\tabcolsep{5pt} 
  \begin{tabular}{lcccccc}
    \toprule
    \textbf{Agent Identity} &
    $\mathcal{C}$ & $\Delta\mathcal{C}$ &
    $\mathcal{D}$ & $\Delta\mathcal{D}$ &
    $\mathcal{T}$ & $\Delta\mathcal{T}$ \\
    \midrule
    \rowcolor{gray!15}
    \textsc{No Identities (Baseline)} & $82.13 \pm 18.96$ & -- & $0.400\pm0.017$ & -- & $0.308\pm0.018$ & -- \\
    \textsc{Cardiac Surgeon} & $\textbf{98.41}\pm\textbf{4.76}$ & \textbf{+16.28} & $0.395\pm0.019$ & \underline{-0.005} & $0.318\pm0.027$ & +0.010 \\
    \textsc{Qualitative Researcher} & $97.56\pm3.34$ & \underline{+15.43} & $0.397\pm0.014$ & \textbf{-0.003} & $0.324\pm0.023$ & +0.016 \\
    \textsc{Medical Doctor} & \underline{$96.83\pm8.98$} & +14.70 & $0.389\pm0.025$ & -0.011 & $0.334\pm0.007$ & \underline{+0.026} \\
    \textsc{Psychologist} & $93.67\pm2.35$ & +11.54 & $0.359\pm0.018$ & -0.041 & $0.325\pm0.015$ & \textbf{+0.027} \\
    \bottomrule
  \end{tabular}
  \caption{\textbf{Performance of Identity-Augmented Agents} Evaluation results across core metrics: credibility ($\mathcal{C}$), dependability ($\mathcal{D}$), and transferability ($\mathcal{T}$). 
  Values denote the mean and standard deviation computed over nine transcripts.  Bolded values indicate the best performance, and underlined values denote the second-best. 
  $\Delta$ columns indicate improvements over the baseline. 
  Higher values are better for $\mathcal{C}$, $\mathcal{D}$, and $\mathcal{T}$. Definitions for each evaluation metric are in Section~\ref{sec:eval-metrics}.}
  \label{tab:results}
\end{table*}

\begin{table*}[t!]
  \centering
  \small
  \setlength\tabcolsep{5pt}
  \begin{tabular}{lcccccc}
    \toprule
    \textbf{Agent Identity} &
    Cosine ($C_{\text{bi}}$) & $\Delta C$ &
    Levenshtein ($D_L$) & $\Delta D_L$ &
    BLEU ($B$) & $\Delta B$ \\
    \midrule
    \rowcolor{gray!15}
    \textsc{No Identities (Baseline)} & 
    $0.132 \pm 0.027$ & -- &
    $0.301 \pm 0.027$ & -- &
    $0.019 \pm 0.008$ & -- \\
    
    \textsc{Cardiac Surgeon} & 
    $0.115 \pm 0.053$ & $-0.017$ & 
    $0.259 \pm 0.089$ & $-0.042$ & 
    $0.020 \pm 0.009$ & $+0.001$ \\

    \textsc{Qualitative Researcher} & 
    $0.107 \pm 0.046$ & $-0.025$ &
    $0.252 \pm 0.082$ & $-0.049$ &
    $0.014 \pm 0.007$ & $-0.005$ \\

    \textsc{Medical Doctor} & 
    $0.112 \pm 0.018$ & $-0.020$ &
    $0.287 \pm 0.025$ & $-0.014$ &
    $0.018 \pm 0.006$ & $-0.001$ \\

    \textsc{Psychologist} & 
    $0.121 \pm 0.029$ & $-0.011$ &
    $0.282 \pm 0.021$ & $-0.019$ &
    $0.020 \pm 0.006$ & $+0.001$ \\
    \bottomrule
  \end{tabular}

  \caption{\textbf{Semantic and Lexical Alignment Metrics.} Mean ± standard deviation scores across three metrics used to evaluate alignment between LLM-generated and human-generated themes: bidirectional Cosine Similarity ($C_{\text{bi}}$), Levenshtein Distance ($D_L$), and BLEU Score ($B$). $\Delta$ columns represent absolute changes from the baseline (No Identities). Higher values indicate better alignment, except for $D_L$, where lower values indicate better alignment.}
  \label{tab:alignment-metrics}
\end{table*}

\begin{table*}[t!]
  \centering
  \footnotesize
  \begin{tabular}{clp{7.4cm}}
    \toprule
    \textbf{\#} & \textbf{Human Theme} & \textbf{Closest LLM Themes (\texttt{iteration=5})} \\
    \midrule
    1 & Clarity of potential risks and outcomes & Desiring Comprehensive Data on Long-Term Outcomes \\
      & & Seeking Clarity and Understanding about My Child's \mbox{Condition} \\
    2 & Freedom from hypervigilance related to the condition & Feeling Overwhelmed by Emotional Uncertainty \\
      & & Living with Constant Anxiety about Child's Health \\
    3 & The diagnosis given in a compassionate \mbox{and empathic way} & Experiencing Relief from Diagnosis and Support \\
      & & Feeling Overwhelmed by Medical Decisions \\
    4 & A sense of control over the future & Balancing Relief and Anxiety about the Future \\
      & & Seeking Reassurance and Clear Communication from \mbox{Medical} Professionals \\
    5 & Being heard and taken seriously by clinicians & Seeking Reassurance and Clear Communication from \mbox{Medical} Professionals \\
      & & Feeling Isolated During the Medical Journey \\
    6 & Individualized support for management decision-making & Advocating for Child's Medical Needs \\
      & & Navigating Surgical Decision Anxiety \\
    7 & Receiving support from others & Finding Strength in Family and Community Support \\
      & & Finding Joy in Family Connections \\
    8 & Being appropriately informed & Seeking Reassurance and Clear Communication from \mbox{Medical} Professionals \\
      & & Desiring Clear Communication from \mbox{Medical} Professionals \\
    9 & Partnership with the care team & Seeking Reassurance and Clear Communication from \mbox{Medical} Professionals \\
      & & Advocating for Child's Medical Needs \\
    10 & Feeling that my child is safe & Coping with Health Crisis Trauma \\
       & & Feeling Overwhelmed by Medical Decisions \\
    11 & Not feeling responsible for the diagnosis and its timing & Struggling with Feelings of Guilt and Responsibility \\
       & & Desiring Proactive Healthcare Measures to Prevent Crises \\
    12 & Appropriately coping with stress, anxiety and depression & Feeling Overwhelmed by Emotional Uncertainty \\
       & & Managing Ongoing Anxiety about My Child's Health \\
    \bottomrule
  \end{tabular}
  \caption{\textbf{Examples of semantic alignment between human and LLM-generated themes.} Each human-generated theme is matched with the two LLM-generated themes exhibiting the highest cosine similarity scores, based on the similarity matrix heatmap (Figure~\ref{fig:heatmap5}) after five refinement iterations incorporating feedback from the feedback agent. These LLM-generated themes are drawn from agents with diverse identities in the \texttt{Auto-TA} pipeline. Although traditional alignment scores are modest, the most similar LLM themes often reflect comparable underlying meanings, and in some cases, elaborate or extend the human-coded interpretations.}
  \label{tab:theme-alignment-updated}
\end{table*}

We evaluated the alignment between LLM-generated and human-generated themes using  $C_{\text{bi}}$, $D_L$, and $B$ (Table~\ref{tab:alignment-metrics}). 
Higher alignment for the Cardiac Surgeon agent likely reflects the presence of a cardiac expert in the human analyst group. Alignment appears influenced by annotator expertise; different expert compositions (e.g., more qualitative researchers) could shift alignment patterns. Even when surface-level overlap is low, identity-augmented agents often capture subtle or overlooked aspects of the data not fully represented in the human-generated themes. This suggests that identity conditioning can expand the thematic space in meaningful and complementary ways.

This observation also points to the limitations of traditional evaluation metrics that rely on surface-level comparison between human and LLM-generated themes. As shown in Table~\ref{tab:theme-alignment-updated}, low scores on metrics such as $B$ do not necessarily imply incorrect themes. Rather, they underscore the need for evaluation criteria better suited to the interpretive nature of qualitative research, where multiple valid versions of TA can coexist. 

Our preliminary results showed no significant changes using the four criteria from previous studies, including distinctiveness and actionability (Figure~\ref{fig:reviewer-feedback}). This may be due to the use of a coarse 5-point integer scoring system within a form-filling paradigm using LLM evaluators, which may lack the resolution to capture micro-scale improvements. To address this, we propose adopting a scalar scoring system (e.g., continuous values between 1.00 and 5.00) to better reflect incremental changes in theme quality during iterative refinement.

\section{Limitations}

A key limitation of our approach lies in the assumption that alignment with human-generated themes is a sufficient indicator of thematic quality. In practice, there may be multiple valid sets of \textit{right} themes for the same transcript, depending on the perspective and interpretive lens of the analyst. As such, high alignment with a human-coded reference set does not necessarily imply better or more meaningful TA. 
This observation aligns with findings from prior work. For instance, a cardiac expert cited in ~\citet{mery2023journey} noted that approximately two-thirds of themes tend to be straightforward and expected, regardless of the analyst’s background. However, the remaining one-third are more variable and dependent on individual expertise and interpretation. 
Beyond this conceptual limitation, several practical constraints remain. The framework has not yet been tested across different domains, leaving its ability to generalize to other clinical or non-clinical settings uncertain. In its current form, the architecture does not support interaction among agents, which limits opportunities for collaborative reasoning or negotiated theme development. Evaluation is also limited to a single comparison with human-coded themes, offering little insight into the stability of outputs across multiple runs. Moreover, the system shows notable sensitivity to prompt wording, where minor variations can lead to substantially different thematic results and may raise concerns about reproducibility in real-world applications.

\section{Conclusion and Future Work}

We presented \texttt{Auto-TA}, a fully automated, multi-agent LLM framework for end-to-end thematic analysis of unstructured clinical narratives. Role-conditioned agents enhanced alignment with human-coded themes, particularly when identities matched annotator expertise. Despite low traditional metric scores, identity-augmented agents captured valid insights and expanded the thematic space. Auto-TA demonstrates potential for scalable, expert-informed qualitative analysis, supporting future extensions through domain adaptation, reinforcement learning, and agent collaboration.

Future work should explore evaluation methods that go beyond surface-level alignment, focusing on the distinctiveness, utility, and interpretive depth of LLM-generated themes in specific clinical or research contexts. Bridging machine efficiency and human interpretability will likely require iterative collaboration with domain experts. Incorporating expert feedback into a reinforcement learning framework could enable adaptive refinement based on human judgment. Expanding the framework to new domains will help assess its generalizability, while structured dialogue or negotiation among agents may better mirror human qualitative analysis. An identity generation agent that selects optimal personas based on dataset context could further improve relevance. Finally, examining reproducibility and sensitivity to prompt variation is essential for real-world reliability.

\clearpage
\bibliography{custom}

\section*{Appendix}
\appendix
\section{Dataset Details}
\subsection{Original Corpus Composition and Participant Summary}
\label{appendix:corpus}

Between February and September 2020, transcripts were collected from 19 moderated focus groups (Experience Groups, 90 minutes, 3–7 participants), 35 semi-structured 1:1 interviews (60 minutes), and 4 co-design workshops aimed at validating emerging journey maps. The full composition of the parent corpus is shown in Table~\ref{tab:corpus}.

\begin{table}[htbp]
    \centering
    \footnotesize
    \setlength{\tabcolsep}{4pt}
    \caption{Parent corpus composition}
    \small
    \label{tab:corpus}
    \begin{tabular}{lccc}
        \toprule
        & \# Sessions & Participants & Approx.\ Words \\
        \midrule
        EG Sessions & 19 & 134 & $\sim$260\,k \\
        1:1 Interviews & 35 & 56$^\dagger$ & $\sim$210\,k \\
        Workshops & 4 & 58 & $\sim$50\,k \\
        \midrule
        \textbf{Total} & 58 & 170 & $\sim$520\,k \\
        \bottomrule
    \end{tabular}
    \vspace{2pt}
    {\footnotesize $^\dagger$35 family participants + 21 stakeholders}
\end{table}

Of the 96 survey respondents, most were parents (n=52) or patients (n=29), with smaller numbers of siblings (n=9), partners (n=4), and fetal case participants (n=4). Females comprised the majority across groups, especially among parents (77\%) and patients (55\%). Most respondents identified as White (81\% of parents, 78\% of patients), with smaller proportions identifying as Hispanic/Latino or Black.

\subsection{Quote Identification and Traceability}

Each statement in the transcripts is tagged with a unique Quote ID to support reference and traceability during TA. The format \texttt{[P\#\_S\#\#\#]} identifies both the speaker and the sequence: \texttt{P\#} denotes the participant (e.g., \texttt{P1} = Participant 1), and \texttt{S\#\#\#} marks the specific utterance from that participant. This system facilitates coding accuracy and enables cross-referencing between annotations and original context. A total of 85 participants contributed to the subset analyzed in this study.

\subsection{Visualizing Transcript Embeddings via t-SNE}

\begin{figure}[H]
  \centering
  \includegraphics[width=0.5\textwidth]{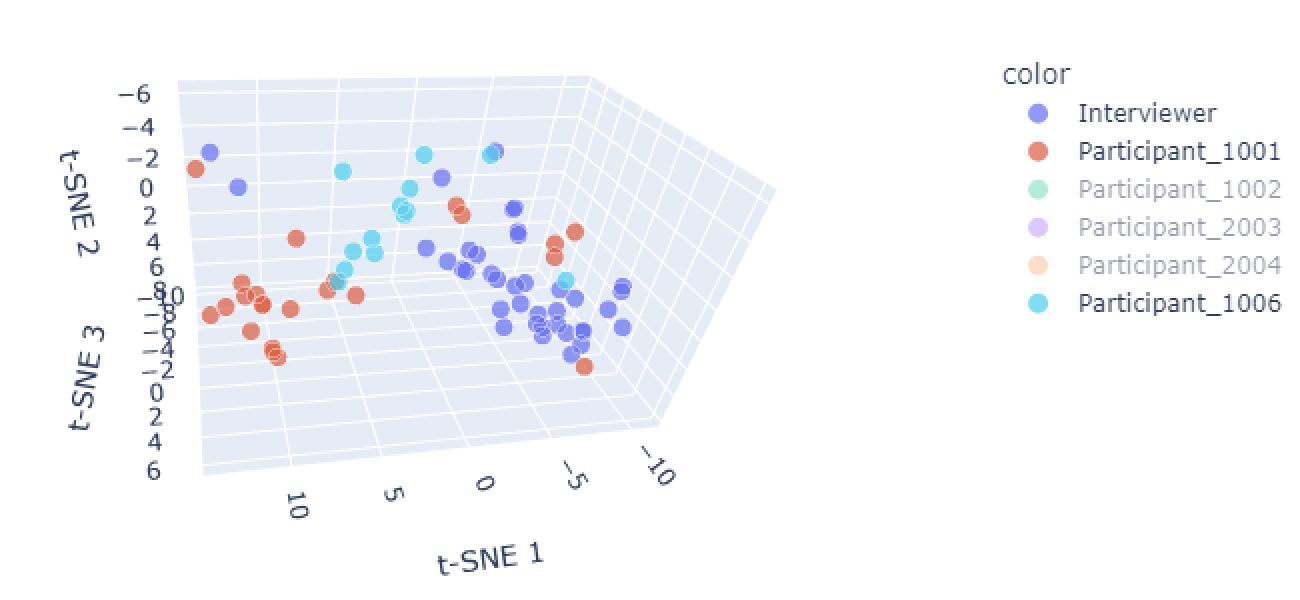}
  \caption{3D t-SNE projection of sentence embeddings from Interview 1, showing clusters by speaker identity.}
  \label{fig:tsne3d}

  \vspace{1em} 

  \includegraphics[width=0.50\textwidth]{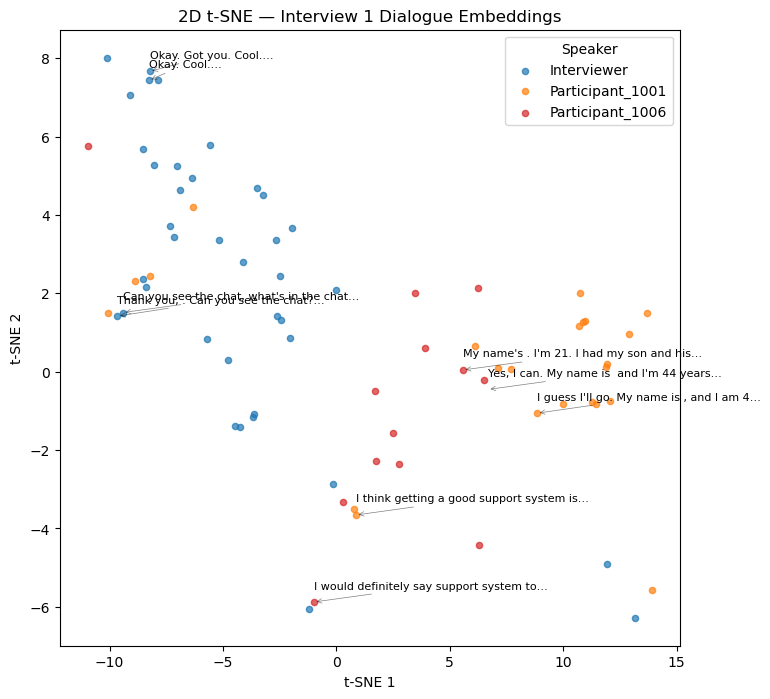}
  \caption{2D t-SNE projection of sentence embeddings with representative labels and speaker roles.}
  \label{fig:tsne2d}
\end{figure}

\subsection{Ethics Statement}

The original study was approved by the UT Austin Institutional Review Board [Protocol \#2019080031] and registered on \textsc{ClinicalTrials.gov} [NCT04613934]. The dataset was generated in the study by \citet{mery2023journey}.

\section{Additional Abbreviations}
\label{sec:appendix_abbr}

Table~\ref{tab:glossary} lists abbreviations that appear frequently in the paper for ease of reference.

\begin{table}[ht!]
\centering
\small
\setlength{\tabcolsep}{4pt}
\begin{minipage}{0.5\textwidth} 
\renewcommand{\arraystretch}{1.3}
\begin{tabular}{@{}p{2cm}p{5.5cm}@{}}
\toprule
\textbf{Term} & \textbf{Definition} \\
\midrule
\textbf{Agent} & \RaggedRight An autonomous computational entity (LLM) that interacts with other agents or the environment to perform specific tasks. \\
\textbf{CHD} & Congenital Heart Disease \\
\textbf{SV-CHD} & Single-Ventricle Congenital Heart Disease \\
\textbf{AAOCA} & \RaggedRight A congenital heart condition where a coronary artery arises from the aorta in an atypical location. \\
\textbf{TA} & \RaggedRight Thematic analysis, a method for identifying and reporting patterns, as proposed by~\citet{braun2006using}. \\
\textbf{LLM} & Large Language Model \\
\textbf{RL} & Reinforcement Learning \\
\textbf{RLHF} & Reinforcement Learning from Human Feedback \\
\textbf{PPO} & Proximal Policy Optimization~\citep{Schulman2017PPO} \\
\textbf{IRB} & Institutional Review Board \\
\bottomrule
\end{tabular}
\caption{Glossary of key terms and abbreviations.}
\label{tab:glossary}
\end{minipage}
\end{table}

\section{Evaluation Metrics Details}

\label{app:evaluation-metrics}
\paragraph{Levenshtein Distance ($D_L$)} To capture lexical similarity between themes, we compute the normalized Levenshtein distance between each pair \((t_i, \ell_j)\), defined as the minimum number of single-character edits (insertions, deletions, or substitutions) required to transform one string into the other.  
The normalized form is given by:
\[
\text{Levenshtein}(t_i, \ell_j) = \frac{\text{edit\_distance}(t_i, \ell_j)}{\max(|t_i|, |\ell_j|)},
\]
where \(|t_i|\) and \(|\ell_j|\) denote the lengths of the respective strings.  
For interpretability, we convert distance into a similarity score:
\[
\text{sim}_{\text{lev}}(t_i, \ell_j) = 1 - \text{Levenshtein}(t_i, \ell_j),
\]
so that values closer to 1 indicate higher surface-level string similarity.  
We report the average maximum similarity per human theme:
\[
D_L = \frac{1}{n} \sum_{i=1}^{n} \max_j \text{sim}_{\text{lev}}(t_i, \ell_j),
\]
which assesses how well each human theme is lexically approximated by its most similar LLM-generated theme.

\section{\texorpdfstring{Dependability ($\mathcal{D}$) and Transferability ($\mathcal{T}$) Score Details}{Dependability (D) and Transferability (T) Score Details}}

Table~\ref{tab:rouge-results} presents detailed ROUGE-1 and ROUGE-2 scores used to assess theme dependability ($\mathcal{D}$) and transferability ($\mathcal{T}$) across agent identities.

\begin{table*}[h!]
  \centering
  \scriptsize
  \setlength{\tabcolsep}{4pt}
  \begin{tabular}{lcccccccccc}
    \toprule
    \textbf{Agent Identity} &
    $\mathcal{R}_1$ & $\Delta\mathcal{R}_1$ &
    $\mathcal{R}_2$ & $\Delta\mathcal{R}_2$ &
    $\mathcal{R}'_1$ & $\Delta\mathcal{R}'_1$ &
    $\mathcal{R}'_2$ & $\Delta\mathcal{R}'_2$ \\
    \midrule
    \rowcolor{gray!15}
    \textsc{No Identities (Baseline)} & $0.564 \pm 0.020$ & -- & $0.236 \pm 0.018$ & -- & $0.437 \pm 0.020$ & -- & $0.180 \pm 0.018$ & -- \\
    \textsc{Cardiac Surgeon} & $0.563 \pm 0.022$ & $-0.001$ & $0.227 \pm 0.021$ & $-0.009$ & $0.449 \pm 0.028$ & $+0.012$ & $0.188 \pm 0.026$ & $+0.008$ \\
    \textsc{Qualitative Researcher} & $0.538 \pm 0.020$ & $-0.026$ & $0.255 \pm 0.016$ & $+0.019$ & $0.456 \pm 0.030$ & $+0.019$ & $0.192 \pm 0.022$ & $+0.012$ \\
    \textsc{Medical Doctor} & $0.546 \pm 0.028$ & $-0.018$ & $0.232 \pm 0.026$ & $-0.004$ & $0.464 \pm 0.009$ & $+0.027$ & $0.204 \pm 0.006$ & $+0.024$ \\
    \textsc{Psychologist} & $0.488 \pm 0.026$ & $-0.076$ & $0.229 \pm 0.018$ & $-0.007$ & $0.452 \pm 0.020$ & $+0.015$ & $0.198 \pm 0.016$ & $+0.018$ \\
    \bottomrule
  \end{tabular}
  \caption{
Lexical overlap scores used in evaluating dependability and transferability. 
$\mathcal{R}_1$ and $\mathcal{R}_2$ denote bidirectional ROUGE-1 and ROUGE-2 scores, respectively, computed across two independent theme generation runs from the same transcript (see Eq.~\ref{eq:r1}--\ref{eq:r2}); their average defines dependability $\mathcal{D}$ (Eq.~\ref{eq:dependability}). 
$\mathcal{R}'_1$ and $\mathcal{R}'_2$ denote analogous ROUGE scores computed between themes generated from different transcript subsets (train vs. validation) and define transferability $\mathcal{T}$ (Eq.~\ref{eq:transferability}). 
Higher values indicate stronger consistency or generalizability.}
  \label{tab:rouge-results}
\end{table*}

\section{Supplementary Results on Theme Alignment}

\begin{figure}[h!]
\centering
\begin{subfigure}[t]{0.48\textwidth}
    \centering
    \includegraphics[width=\linewidth]{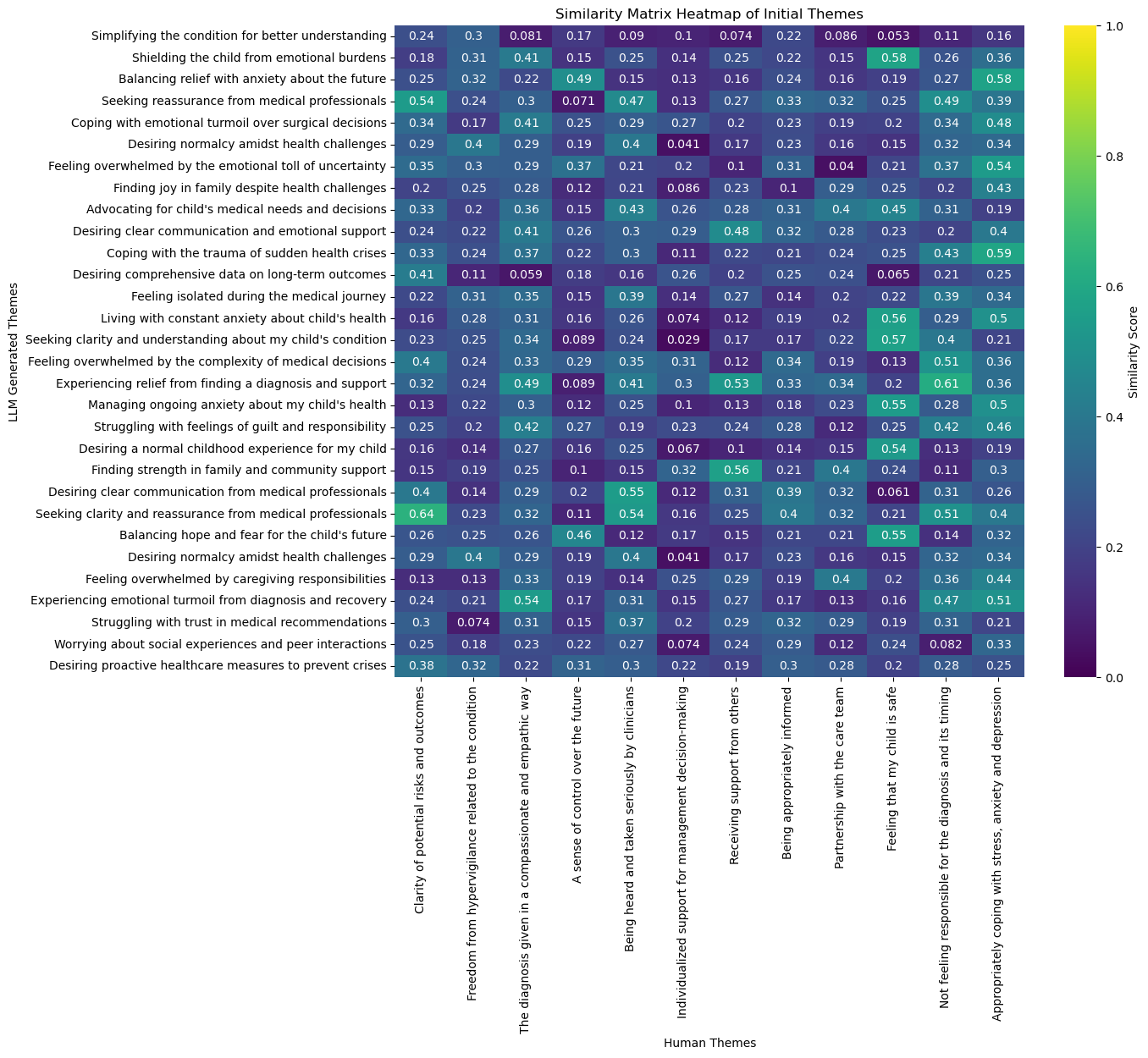}
    \caption{Cosine Similarity at Iteration 0}
    \label{fig:heatmap0}
\end{subfigure}
\hfill
\begin{subfigure}[t]{0.48\textwidth}
    \centering
    \includegraphics[width=\linewidth]{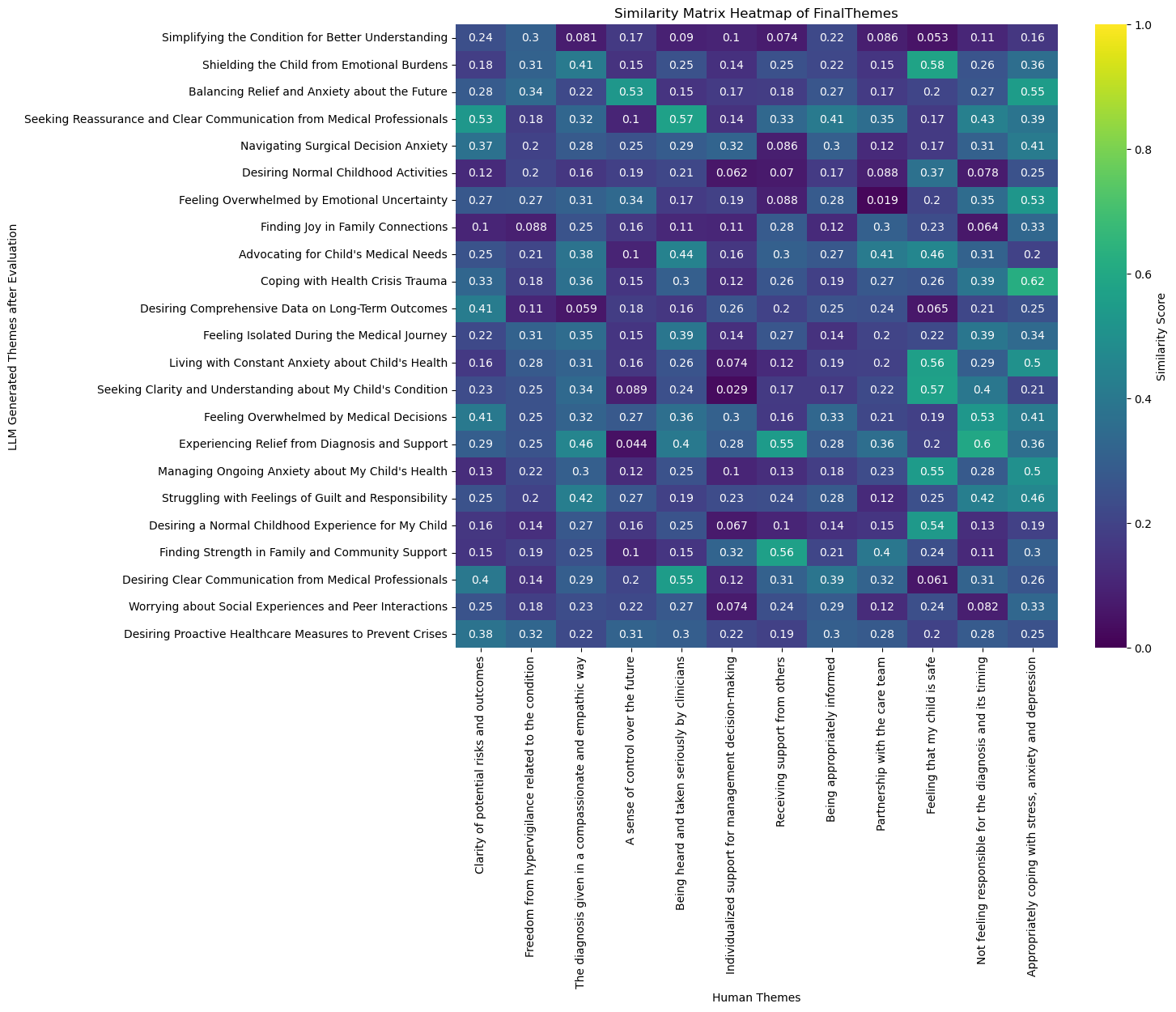}
    \caption{Cosine Similarity at Iteration 5}
    \label{fig:heatmap5}
\end{subfigure}
\caption{Comparison of LLM-human theme similarity across iterations. Higher similarity scores in later iterations suggest better thematic alignment after feedback-driven refinement.}
\label{fig:heatmap-comparison}
\end{figure}

\begin{figure}[h!]
\centering
\includegraphics[width=0.5\textwidth]{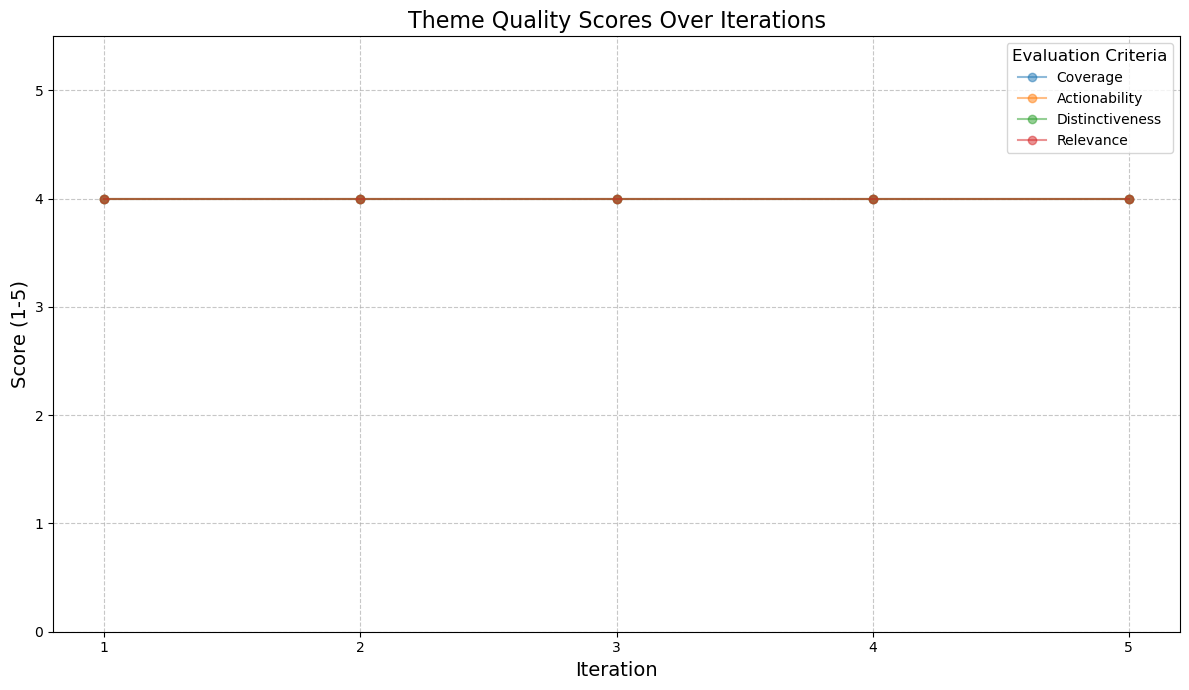}
\caption{Theme quality scores (Coverage, Actionability, Distinctiveness, and Relevance) across refinement iterations. Flat trends suggest that integer-based evaluation lacks granularity to detect micro-level improvements.}
\label{fig:reviewer-feedback}
\end{figure}

This section provides supplementary visualizations related to theme alignment with human ground truth. Figure~\ref{fig:heatmap-comparison} shows improved cosine similarity between LLM and human themes from iteration 0 to 5. Figure~\ref{fig:reviewer-feedback} indicates that reviewer-assigned scores remained static, highlighting the need for more sensitive metrics. Table~\ref{tab:theme-alignment-concise} shows semantically aligned theme pairs from the \textsc{Cardiac Surgeon} agent, highlighting shared or complementary meanings despite low similarity scores.

\begin{table*}[t!]
  \centering
  \small
  \setlength{\tabcolsep}{4pt}
  \begin{tabular}{clp{7.5cm}}
    \toprule
    \textbf{\#} & \textbf{Human Theme} & \textbf{Closest LLM Themes (Cardiac Surgeon, \footnotesize{\texttt{iteration=0}})} \\
    \midrule
    1 & Clarity of potential risks and outcomes & Postoperative Outcomes and Recovery  \\
    & & Concerns About Long-Term Outcomes and Data Scarcity \\
    2 & Freedom from hypervigilance related to the condition & majority of the mental health affect from PTSD \\
    & & Initial Underestimation of Condition Severity\\ 
    3 & The diagnosis given in a compassionate and empathic way & Relief in Diagnosis; Initial Diagnosis and Confusion \\
    4 & A sense of control over the future & Post-Surgery Clearance and Future Concerns\\
    5 & Being heard and taken seriously by clinicians & Privacy and Protection of Child's Emotional Well-being  \\
    6 & Individualized support for management decision-making & Decision for Surgery; Decision-Making and Surgical Considerations \\
    7 & Receiving support from others & Role of Support Systems \\
    8 & Being appropriately informed & Privacy and Protection of Child's Emotional Well-being \\
    9 & Partnership with the care team & the kids on his team."; Trust in Medical Team \\
    10 & Feeling that my child is safe & Navigating Parent-Child Communication \\
    11 & Not feeling responsible for the diagnosis and its timing & Diagnosis and Initial Reactions \\
    12 & Appropriately coping with stress, anxiety and depression & Depression; Anxiety \\
    \bottomrule
  \end{tabular}
  \caption{\textbf{Examples of semantic alignment between human and LLM-generated themes.} Each human-generated theme is paired with two closely aligned LLM-generated themes from the \textsc{Cardiac Surgeon} identity. Although traditional alignment scores between human- and LLM-generated themes are low, many theme pairs share similar underlying meanings, and some even build upon or encompass each other.}
  \label{tab:theme-alignment-concise}
\end{table*}

\section{Example Prompt with Identities}
\label{app:prompt-iden}
\begin{tcolorbox}[colback=gray!5!white, colframe=gray!50!black, title=Prompt for Surgical Coder Identity]
\small
You are an experienced cardiac surgeon specializing in Anomalous Aortic Origins of Coronary Artery (AAOCA).  
Your role is to code patient narratives from a surgical perspective.

Our end goal is to perform TA on the provided transcript~\citep{braun2006using}.

The steps that should be performed are as follows:

\begin{enumerate}
  \item \textbf{Familiarization} \\
  Read and re-read the data to become deeply familiar with it.
  
  \item \textbf{Generating Initial Codes} \\
  Systematically code interesting features across the dataset.

  \item \textbf{Searching for Themes} \\
  Group codes into potential themes, collating relevant data.

  \item \textbf{Reviewing Themes} \\
  Check if themes work in relation to coded extracts and the full dataset.

  \item \textbf{Defining and Naming Themes} \\
  Refine each theme and define its essence and scope.

  \item \textbf{Producing the Report} \\
  Final analysis and write-up with evidence-rich examples.
\end{enumerate}

Your job is to perform \textbf{ONLY Step 2}.

You \textbf{MUST} provide the unique Quote IDs and descriptions about the codes according to the context of the original text.  
The Quote IDs in the transcript are marked as \texttt{[P1\_S002]} for example.

Here is the Original transcript:
\end{tcolorbox}

\section{Example Prompt and Generated Codes}
\subsection{Example Prompt for Code Generation}

\begin{tcolorbox}[
  colback=gray!5!white,
  colframe=black!75,
  title=Prompt: Surgical Perspective on Coding Patient Narratives,
  breakable,
  enhanced,
  arc=4pt,          
  before skip=10pt,
  after skip=10pt
]
\small
You are an experienced cardiac surgeon specializing in Anomalous Aortic Origins of Coronary Artery (AAOCA).  
Your role is to code patient narratives from a surgical perspective.

Our end goal is to perform TA on the provided transcript~\citep{braun2006using}.

The steps that should be performed are as follows:
\begin{enumerate}[leftmargin=*]
  \item \textbf{Familiarization:} Read and re-read the data to become deeply familiar with it.
  \item \textbf{Generating Initial Codes:} Systematically code interesting features across the dataset.
  \item \textbf{Searching for Themes:} Group codes into potential themes, collating relevant data.
  \item \textbf{Reviewing Themes:} Check if themes work in relation to coded extracts and the full dataset.
  \item \textbf{Defining and Naming Themes:} Refine each theme and define its essence and scope.
  \item \textbf{Producing the Report:} Final analysis and write-up with evidence-rich examples.
\end{enumerate}

\textbf{Your task is to perform ONLY Step 2.}  
You \textbf{MUST} include unique Quote IDs (e.g., \texttt{[P1\_S002]}) with descriptions grounded in the transcript.
\end{tcolorbox}

\subsection{Example of Generated Codes}
\label{app:example-codes}

Note that Vanilla refers to agents with no designated identities.

\begin{tcolorbox}[
  colback=white,
  colframe=gray!60!black,
  title=Initial Codes from Surgeon Perspective,
  breakable,
  enhanced,
  arc=4pt,              
  before skip=10pt,
  after skip=10pt
]
\small
\begin{enumerate}[leftmargin=*]
  \item \textbf{Understanding AAOCA}  
    \textit{Quote ID: [P1\_S001]} \\
    Use of acronyms and terminology (AAOCA, ALCA) reflects familiarity with medical vocabulary.

  \item \textbf{Variation in Medical Conditions}  
    \textit{Quote ID: [P2\_S001]} \\
    Illustrates how the anomaly presents differently depending on which coronary artery is affected.

  \item \textbf{Simplifying Medical Terminology}  
    \textit{Quote ID: [P4\_S001]} \\
    Parents often simplify terminology for ease of communication with others.

  \item \textbf{Lack of Understanding Among Non-Medical Individuals}  
    \textit{Quote ID: [P4\_S003]} \\
    Indicates that detailed terminology is typically only understood by healthcare professionals.

  \item \textbf{Emotional Impact of Medical Complexity}  
    \textit{Quote ID: [P5\_S001]} \\
    Parents express emotional burden due to difficulty explaining the condition.

  \item \textbf{Privacy and Child’s Emotional Well-being}  
    \textit{Quote ID: [P2\_S003]} \\
    Parents shield children from information to avoid psychological stress.

  \item[]
  \centering\textbf{\dots}
\end{enumerate}
\end{tcolorbox}

\begin{tcolorbox}[
  colback=white,
  colframe=gray!60!black,
  title=Initial Codes from Vanilla Identity,
  breakable,
  enhanced,
  arc=4pt,              
  before skip=10pt,
  after skip=10pt
]
\small
\begin{enumerate}[leftmargin=*]
  \item \textbf{Use of Acronyms and Simplification}  
  \textit{Quote ID: [P1\_S001], [P2\_S001], [P4\_S001], [P5\_S001], [P3\_S001]} \\
  Participants refer to the condition using acronyms like “AAOCA” and simplified terms to enhance comprehension.

  \item \textbf{Emotional Impact and Coping}  
  \textit{Quote ID: [P5\_S008], [P2\_S007], [P4\_S007], [P1\_S002]} \\
  Expressed emotions include devastation, stress, and relief; a recurring theme across interviews.

  \item \textbf{Lack of Information and Data}  
  \textit{Quote ID: [P4\_S010], [P5\_S017], [P1\_S013]} \\
  Participants highlight frustrations stemming from the rarity of AAOCA and limited clinical data.

  \item \textbf{Decision-Making Challenges}  
  \textit{Quote ID: [P4\_S008], [P2\_S011], [P5\_S021]} \\
  Ambiguity in choosing surgery vs. watchful waiting due to mixed medical recommendations.

  \item \textbf{Protective Parenting and Monitoring}  
  \textit{Quote ID: [P2\_S014], [P3\_S012], [P5\_S024]} \\
  Parents remain highly vigilant, often suppressing their concerns to avoid alarming children.


  \item[]
  \centering\textbf{\dots}



  \item \textbf{Sharing Experiences and Offering Help}  
  \textit{Quote ID: [P5\_S027], [P1\_S021]} \\
  Participants value peer mentorship and express willingness to support newly diagnosed families.

  \item \textbf{Impact on Daily Life and Activities}  
  \textit{Quote ID: [P3\_S004], [P4\_S012], [P5\_S015]} \\
  The condition disrupts routines and limits children's participation in physical activities.
\end{enumerate}
\end{tcolorbox}

\end{document}